# Small Object Detection for Indoor Assistance to the Blind using YOLO NAS Small and Super Gradients

Mrs.Rashmi BN
Assistant Professor ,CSE,
JSS Academy of Technical Education
Bengaluru

Dr. R.Guru ,
Associate Professor,CSE,
SJCE,Mysore

Dr.Anusuya M A
Associate Professor,CSE
SJCE,Mysore

## Abstract

Advancements in object detection algorithms have opened new avenues for assistive technologies that cater to the needs of visually impaired individuals. This paper presents a novel approach for indoor assistance to the blind by addressing the challenge of small object detection. We propose a technique YOLO NAS Small architecture, a lightweight and efficient object detection model, optimized using the Super Gradients training framework. This combination enables real-time detection of small objects crucial for assisting the blind in navigating indoor environments, such as furniture, appliances, and household items. Proposed method emphasizes low latency and high accuracy, enabling timely and informative voice-based guidance to enhance the user's spatial awareness and interaction with their surroundings. The paper details the implementation, experimental results, and discusses the system's effectiveness in providing a practical solution for indoor assistance to the visually impaired.

***Keywords: Small Object Detection, Indoor Navigation, Assistive Technology, YOLO NAS Small, Super Gradients***

## 1. Introduction

Assisting the visually impaired in navigating and interacting with indoor environments presents a significant challenge. Traditional approaches, such as canes and guide dogs, while helpful, have limitations in providing detailed information about the surroundings. Computer vision, particularly object detection, offers a promising avenue to enhance indoor navigation for the blind by providing real-time information about nearby objects. This paper introduces a novel system designed to aid the visually impaired in indoor environments by leveraging the power of small object detection. Our approach combines the efficiency of the YOLO NAS Small architecture with the advanced training capabilities of the Super Gradients framework. This combination allows us to achieve real-time performance crucial for delivering timely voice-based guidance to the user [1].We focus on detecting small objects commonly found in indoor settings, such as furniture, appliances, and household items, which are essential for safe and independent navigation. By accurately identifying and locating these objects, our system can provide blind individuals with a greater understanding of their surroundings, enabling them to make informed decisions about their movements. This paper details our proposed system's architecture, implementation, and experimental evaluation. We demonstrate its effectiveness in detecting small objects in indoor environments and discuss its potential to significantly improve the lives of visually impaired individuals by fostering greater independence and safety in their daily lives [2].

### 1.1 Problem Statement

While object detection shows promise for assisting the visually impaired with indoor navigation, accurately detecting small objects in real-time, crucial for safe and independent movement, remains a significant challenge. Existing object detection models often struggle to balance accuracy and efficiency when dealing with small objects, limiting their practicality for real-time assistive technologies. This research addresses this problem by developing a system that leverages the YOLO NAS Small architecture and Super Gradients training framework to achieve accurate and efficient small object detection for providing real-time, voice-based guidance to the blind in indoor environments.

## 1.2 The Main Contributions made in the work

This research makes significant contributions by:

- Implemented a novel system specifically designed for small object detection in indoor environments to assist the visually impaired, leveraging the efficiency of the YOLO NAS Small architecture and the advanced training capabilities of the Super Gradients framework.
- Demonstrating the system's effectiveness in accurately detecting small objects crucial for indoor navigation, such as furniture, appliances, and household items, while maintaining real-time performance necessary for practical assistive technology.
- Providing a foundation for future research in developing robust and user-friendly assistive technologies for the visually impaired by showcasing the potential of combining state-of-the-art object detection models with tailored training frameworks.

This paper is framed as follows: section 1, provides the brief introduction with problem statement and contributions made in this work. Section 2, provides the related work and limitations of the existing work. Section 3 provides the detailed description of the methodology followed and algorithms proposed. Section 4 presents the results obtained and comparative analysis and finally conclusion is presented in section 5.

## 2. Related Work

Assisting the visually impaired in indoor environments is a crucial challenge that has garnered significant attention in recent years [3]. One promising approach to this problem is the use of computer vision techniques for detecting and recognizing small objects, which can then be conveyed to the user through voice assistance [4]. The YOLO (You Only Look Once) object detection model has emerged as a popular choice for this task due to its real-time performance and ability to handle a wide range of object types. However, the standard YOLO architecture may struggle with the detection of small objects, which are often crucial for assisting the blind in indoor settings. To address this limitation, researchers have proposed the use of the YOLO NAS Small architecture, which is designed to be more efficient and effective at detecting small objects [5]. This model utilizes a neural network-based regression approach to detect objects and a classification algorithm to identify them, streamlining the process into a single, fast pass through the network.

The field of small object detection has witnessed remarkable advancements in recent years, primarily driven by the advent of convolutional neural networks (CNNs). These developments have led to the emergence of two distinct methodologies: anchor-free and anchor-based approaches [18]. Within the anchor-based paradigm, researchers have explored both single-stage and two-stage algorithms. Two-stage algorithms, exemplified by the RCNN family, involve an initial region proposal generation followed by bounding box refinement. While these methods achieve high Mean Average Precision (mAP), their multi-stage architecture often results in slower inference speeds, limiting their applicability in real-time scenarios [19]. Despite efforts to enhance RCNN's efficiency through innovations like Fast-RCNN [20], Faster-RCNN [21], and Mask-RCNN [22], the fundamental two-stage structure continues to pose challenges for real-time performance.

In contrast, single-stage algorithms, notably the You Only Look Once (YOLO) family [23], have garnered significant attention for their speed and efficiency. YOLO's unified network architecture, which simultaneously predicts bounding box coordinates and class probabilities, enables remarkably fast inference. However, early YOLO iterations faced challenges such as relatively lower mAP and difficulties in detecting densely grouped objects. Subsequent YOLO versions have addressed these limitations, culminating in the recent release of YOLOv8. While YOLOv8 represents a significant leap forward in balancing real-time performance and accuracy, there remains potential for further optimization and enhancement in object detection algorithms.

Lin et al. [9] proposed a wearable assistive system utilizing deep learning for object detection and recognition to aid visually impaired individuals. While it doesn't explicitly focus on small object detection, it highlights the potential of deep learning in creating practical assistive technologies. Alagarsamy et al. [10], explored on real-time object detection using YOLO V3 and R-CNN for assisting the visually impaired. It emphasizes the importance of real-time performance and accurate object recognition for providing effective guidance and enhancing independence. Kuriakose et al. [1], presents DeepNAVI, a smartphone-based navigation assistant that utilizes deep learning for obstacle detection and scene recognition. It underscores the need for lightweight models and custom datasets tailored for navigation tasks to assist visually impaired users effectively. Redmon et al. [12], introduces YOLO9000, a significant advancement in the YOLO object detection framework. While not directly focused on assistive technology, it lays the groundwork for faster and more accurate object detection models, which are crucial for real-time applications like assisting the visually impaired.

## 3. Methodology

The initial step in our methodology shown in figure 1, involves capturing images from indoor environments using a device, such as a camera or smartphone, carried by the blind individual. These images serve as the primary input for our system. To ensure the robustness of our detection model, the captured images undergo a comprehensive preprocessing stage. This stage includes image normalization to adjust brightness, contrast, and scale, thereby ensuring consistency in the input data. Additionally, data augmentation techniques, such as rotation, flipping, and scaling, are applied to increase the diversity of the training dataset. This augmentation helps in mitigating overfitting and enhances the model's ability to generalize across different indoor scenarios.The core of our system is the YOLO NAS Small model, specifically designed for efficient small object detection. The model comprises three main components: the backbone network for feature extraction, the neck network for aggregating features at various scales, and the head network for predicting bounding boxes, class probabilities, and confidence scores. To optimize the performance of YOLO NAS Small, we incorporate the Super Gradients optimization technique. This involves efficient gradient computation and the application of advanced training enhancements, such as learning rate scheduling, momentum, and weight decay. These techniques collectively ensure faster convergence and improved accuracy of the model. The training process is conducted on a diverse dataset of indoor images to fine-tune the model parameters for optimal detection performance.

Once the YOLO NAS Small model generates predictions, the outputs undergo a postprocessing stage to refine the detection results. Non-Maximum Suppression (NMS) is employed to eliminate redundant bounding boxes, retaining only the most accurate detections. Further refinement of the bounding boxes is performed to precisely match the detected objects. The final step involves converting the detection results into assistive feedback for the blind individual. This feedback can be delivered through auditory or haptic means, providing real-time information about the presence and location of small objects in the indoor environment. The seamless integration of detection and feedback mechanisms ensures an efficient and user-friendly system for indoor assistance to the blind.

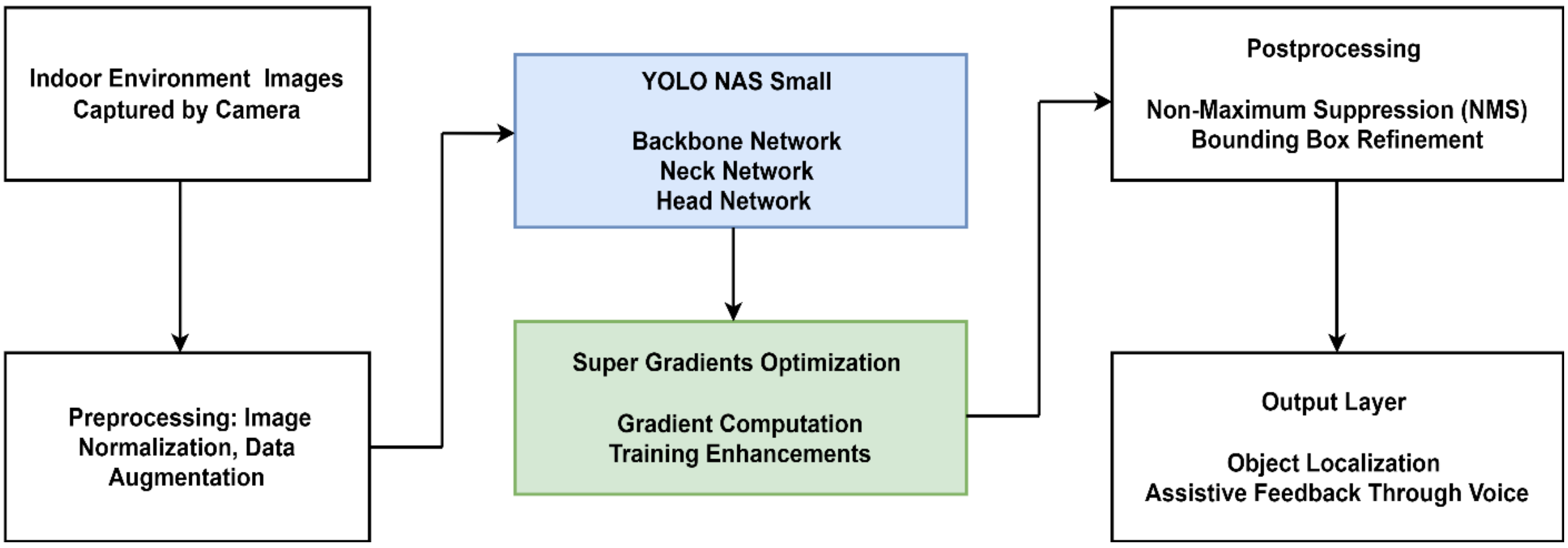

Figure 1: Proposed methodology

The YOLO NAS Small model is particularly well-suited for deployment on resource-constrained devices, such as those used in assistive technologies for the blind [7]. To further enhance the performance of YOLO NAS Small on small objects, we integrate the Super Gradients training framework. Super Gradients is a state-of-the-art deep learning training library that offers advanced optimization techniques and architectural modifications to improve the accuracy and efficiency of neural networks. Our experiments demonstrate the effectiveness of this approach for small object detection in indoor environments, with a focus on objects that are crucial for assisting the blind, such as furniture, appliances, and other household items [8].

### 3.1 Data Collection

To train our system for robust small object detection in indoor environments, we compiled a diverse dataset comprising images captured from various indoor settings, including homes, offices, and public spaces named roboflow. The dataset focused on capturing a wide range of small objects commonly encountered by visually impaired individuals, such as furniture (chairs, tables, shelves), appliances (refrigerators, microwaves, televisions), and household items (books, cups, remote controls). We ensured data diversity by varying lighting conditions, camera angles, and object occlusions to enhance the model's generalization capabilities. The Roboflow is a platform that provides tools and services for building and managing datasets, particularly for computer vision tasks. It offers a wide range of publicly available datasets that can be used for training and testing machine learning models. Roboflow datasets cover a wide range of applications, from object detection and classification to segmentation tasks, and they are used by researchers and developers to build and train computer vision models efficiently [12]. From roboflow platform,

The Yale-CMU-Berkeley (YCB) COCO dataset is a comprehensive and versatile benchmark for object detection and manipulation research [13]. YCB-3 Dataset is utilised for conducting the experiments. YCB-3 is a subset of the YCB (Yale-CMU-Berkeley) Object and Model Set, which is widely used for benchmarking in robotics, computer vision, and machine learning. his dataset combines the rich object variety of the YCB Object and Model Set with the annotation style and evaluation metrics of the popular COCO (Common Objects in Context) dataset. It features a diverse collection of everyday objects, carefully selected to represent a wide range of shapes, sizes, textures, and material properties. The dataset includes high-quality RGB images, depth maps, and precise 6D pose annotations for each object instance. Particularly relevant for our study on small object detection for indoor assistance to the blind, the YCB COCO dataset offers a subset focusing on small objects commonly found in indoor environments. This subset includes items such as food containers, fruits, household items, and sports equipment, making it ideal for training and evaluating models aimed at assisting visually impaired individuals in navigating and interacting with their immediate surroundings. The standardized COCO-style annotations facilitate easy integration with existing object detection frameworks, enabling fair comparisons across different models and methodologies.

## 3.2 Pre-Processing

Pre-processing involved resizing images to a consistent resolution suitable for the YOLO NAS Small architecture while preserving aspect ratios to avoid distortions. We applied data normalization techniques, specifically mean subtraction and standard deviation scaling, to standardize pixel values across the dataset, improving training stability and convergence speed. Data augmentation techniques, including random horizontal flipping, cropping, and minor rotations, were employed to artificially increase the dataset size and expose the model to a wider range of object appearances and orientations, further enhancing its robustness and ability to generalize to unseen scenarios.

## 3.3 YOLO NAS

Deci, a firm that builds production-grade models and tools to construct, optimise, and deploy deep learning models, launched YOLO-NAS [25] in May 2023 as shown in figure 2. With its ability to identify small objects, increase localization precision, and boost performance-per-compute ratio, YOLO-NAS is well suited for real-time edge-device apps. Furthermore, researchers can leverage its open-source architecture. Among YOLO-NAS's innovations are the following:

- Quantization-aware modules (QSP and QCI) [26], which integrate re-parameterization to reduce the accuracy loss during post-training quantization, opt for 8-bit quantization.
- Deci's exclusive NAS technology, AutoNAC, for automatic architecture design.
- Using a hybrid quantization technique, you can quantize specific model components to achieve equilibrium.
- Rather of using normal quantization, which affects all layers, consider latency and accuracy.
- Pre-training consisting of self-distillation, automatically labelled data, and big datasets

After pre-training the model using Objects365 [27], which has two million photos and 365 categories, pseudo-labels were produced using the COCO dataset. Lastly, the original 118k training photos from the COCO dataset are used to train the models. Three YOLO-NAS models with 16-bit precision and FP32, FP16, and INT8 precisions have been made available. They have an AP of 52.2% on MS COCO. YOLO-NAS was made possible by the adaptability of the AutoNAC system, which can handle any work, the details of the data, the establishment of performance targets, and the environment for drawing conclusions. It helps users choose the best structure that provides the ideal balance between accuracy and inference speed for their specific requirements. This technique takes into account the hardware, data, and other components—like quantization and compilers—that are used in the inference process. Furthermore, throughout the NAS process, RepVGG blocks were added to the model architecture to ensure compliance with post-training quantization (PTQ). By changing the depth and locations of the QSP and QCI blocks, they were able to create three different architectures: YOLO-NASS, YOLO-NASM, and YOLO-NASL (S, M, and L for small, medium, and large, respectively).

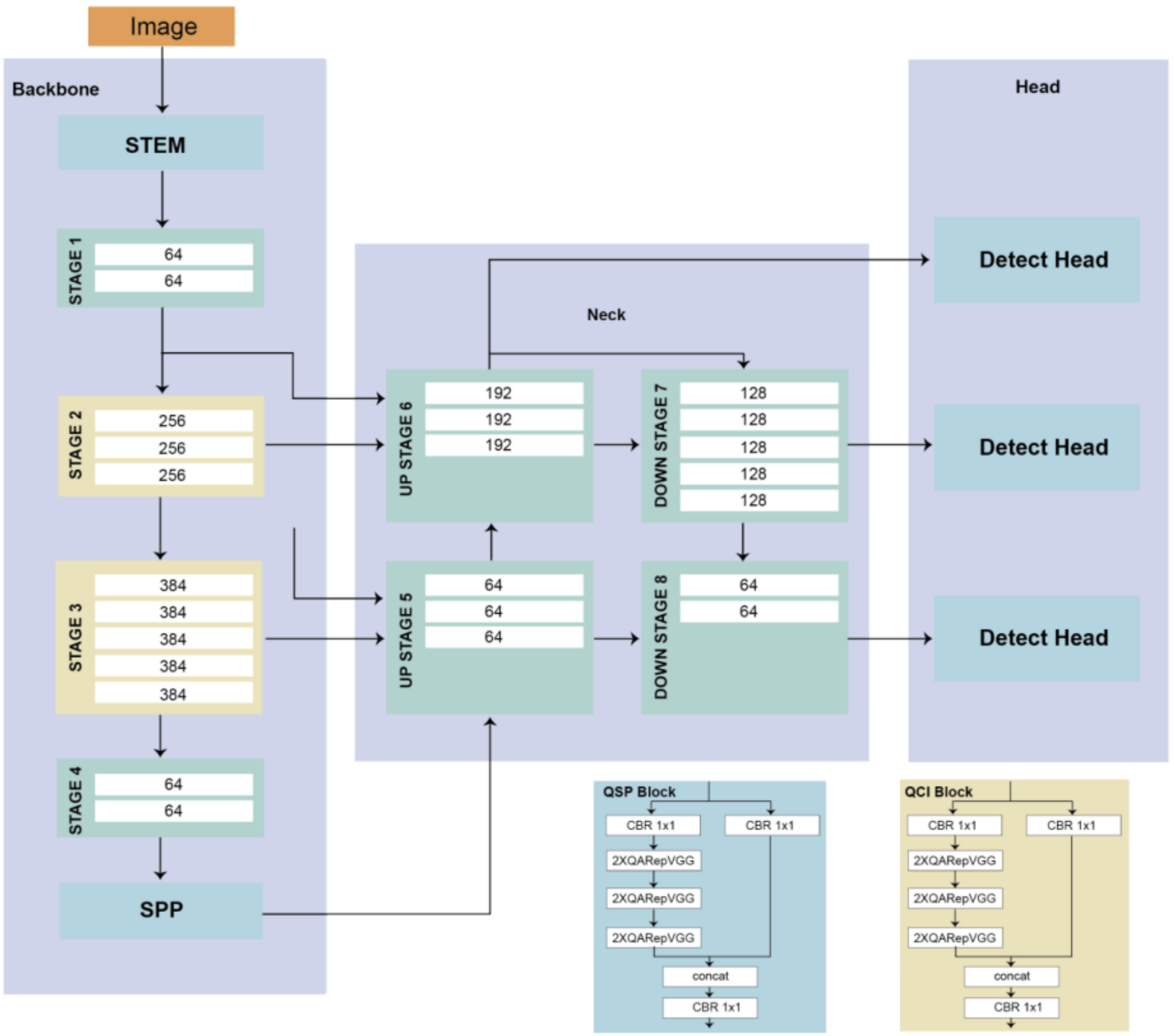

Figure 2: YOLO-NAS architecture. The architecture is found automatically via a neural architecture search (NAS) system called AutoNAC to balance latency vs. throughput [25].

YOLO NAS S, M, and L denote variants of the YOLO architecture that have been optimized through Neural Architecture Search for varying levels of computational resources and performance goals, with the "S", "M", and "L" likely standing for "small", "medium", and "large", respectively. These distinctions typically reflect a trade-off between speed, accuracy, and the computational cost required for training and deploying the models. YOLO NAS S, as the small variant, is designed to operate with fewer computational resources, making it suitable for applications where speed and low latency are critical, such as real-time processing on mobile or edge devices. The architecture of YOLO NAS S is streamlined compared to its larger counterparts; it comprises fewer convolutional layers, filters, and parameters, which results in faster inference times but may slightly compromise the model's accuracy. This variant is particularly useful in scenarios where a quick response is more valuable than pinpoint precision, or where hardware limitations necessitate a lightweight model.

On the other hand, YOLO NAS M and L are progressively larger and more complex versions tailored for systems with more robust computational capabilities. YOLO NAS M finds a middle ground, aiming to balance performance and computational efficiency. It might incorporate more layers and parameters than YOLO NAS S, affording it higher accuracy. YOLO NAS L is the most extensive model of the three, designed for maximum accuracy and precision in scenarios where computational expense is less of a constraint. This could be in scenarios such as cloud computing environments where high-end GPUs are available, and there is a premium on precision for complex tasks like high-resolution video surveillance. Due to their increased size and complexity, YOLO NAS M and L would be expected to have slower inference times but higher accuracy rates compared to YOLO NAS S. YOLO NAS S, M, and L represent different architecture variants derived from the fusion of the YOLO framework with Neural Architecture Search, with the intention of adapting model complexity to match the specifications and constraints of diverse usage scenarios. The 'S', 'M', and 'L' nomenclature typically indicates "Small," "Medium," and "Large," reflecting the scale and complexity of the respective models.

The **YOLO NAS S** architecture is designed for speed and efficiency, tailored to environments with limited computational resources, such as mobile or edge devices. This design prioritizes a smaller network with fewer convolutional layers, fewer filters, and reduced feature map sizes, which results in a lesser number of learnable parameters. The trade-off is often a lower overall accuracy due to the simplified model structure, which restricts its ability to capture complex features. However, YOLO NAS S is optimized for minimal latency, enabling it to perform rapid inferences, which is crucial for real-time applications.In contrast, the **YOLO NAS M** variant strikes a balance between model complexity and computational demand. This medium-sized design involves a more sophisticated architecture than the 'S' variant, with an increased number of layers, filters, and potentially more advanced layer types or connections. While it requires more computational power and memory than YOLO NAS S, it also yields improved accuracy in detecting objects, including text in images and videos, without severely compromising inference speed. It serves a midrange solution, fitting for scenarios where a blend of both performance and efficiency is desired.The largest of the trio, **YOLO NAS L**, is constructed to maximize detection accuracy and is less constrained by computational resource limitations. Featuring the most extensive architecture, YOLO NAS L includes a higher number of convolutional layers with a larger number and variety of filters, greater depth, and possibly additional mechanisms such as attention modules or more complicated branching structures. This complexity allows it to extract detailed features and generalize better over diverse datasets, leading to superior performance on challenging object detection tasks. Nonetheless, this comes at the cost of slower inference times and greater computational demands, rendering it more suitable for high-resource environments such as powerful workstations or cloud-based platforms. These differences are designed to cater to varying processing power and performance requirements:

**YOLO NAS S:**

- **Number of Layers:** Fewer convolutional layers to reduce the depth of the network.
- **Number of Parameters:** Significantly fewer parameters to simplify the model, which helps in reducing the model size and allows for faster inference times.
- **Layer Complexity:** Simplified layers with fewer filters and smaller kernel sizes to minimize computation.
- **Use Case:** Ideal for real-time applications on devices with limited computational resources, such as mobile phones or embedded systems.

**YOLO NAS M:**

- **Number of Layers:** A moderate increase in the number of layers compared to NAS S, adding more depth to the network.
- **Number of Parameters:** More parameters than NAS S but fewer than NAS L, reflecting a mid-range model size and complexity.
- **Layer Complexity:** A balance in layer complexity with a moderate number of filters and potentially varying kernel sizes to enhance feature extraction capabilities without overly increasing computation.
- **Use Case:** Strikes a balance between performance and computational load; suitable for general-purpose applications where there is a need for both accuracy and speed, such as on consumer-grade computers or GPUs.

**YOLO NAS L:**

- **Number of Layers:** The highest number of layers, featuring a deep network architecture.
- **Number of Parameters:** The largest number of parameters representative of a model with high complexity.
- **Layer Complexity:** High complexity layers with a substantial number of filters and potentially larger kernel sizes to maximize the model's capability in capturing nuanced features.
- **Use Case:** Optimized for scenarios where accuracy is the priority and computational resources are abundant, such as on high-grade GPUs in cloud computing environments.

## 3.4 Super Gradients

While the YOLO NAS Small architecture provides an efficient base for our system, achieving optimal performance for small object detection in real-time requires a robust training framework. This is where Super Gradients plays a crucial role. Super Gradients is a powerful deep learning training library that offers advanced optimization algorithms, hyperparameter tuning strategies, and architectural modifications tailored for improved accuracy and efficiency. By integrating Super Gradients into our training pipeline, we leverage these advanced features to fine-tune the YOLO NAS Small model specifically for our task. This includes utilizing optimized learning rate schedules, data augmentation strategies, and loss functions tailored for small object detection. This targeted optimization process ensures that our model effectively learns to identify and localize small objects within indoor environments, ultimately leading to more accurate and reliable real-time assistance for the visually impaired.

## 3.5 Non-Maximum Suppression

Object detection models, including YOLO NAS Small, often predict multiple bounding boxes for a single object, especially small objects, leading to redundant detections. Non-Maximum Suppression (NMS) is a crucial post-processing step that addresses this issue, ensuring that only the most accurate bounding box for each object is retained. In our methodology, NMS analyzes the predicted bounding boxes and their associated confidence scores. It starts by selecting the box with the highest confidence score. Then, it iteratively suppresses boxes that significantly overlap with the selected box and have lower confidence scores. This process continues until only the most confident and non-overlapping boxes, representing distinct object detections, remain. By applying NMS, we refine the model's output, reducing false positives and enhancing the accuracy of small object detection, which is essential for providing reliable guidance to visually impaired users.

---

Algorithm: ***Yolo NAS S + Super gradients + NMS***

---

1 Input: I← input images from the indoor environment, M← YOLO NAS Small model
  $\alpha$ , $\beta$ , $\gamma$ ←hyper parameters for learning rate, momentum and weight decay
  T← training dataset of labelled images with small objects
2 Normalise the input image I
  $I_{norm} = \frac{I-\mu}{\sigma}$, where, $\mu$ is the mean, $\sigma$ is the standard deviation of pixel values
3 Augment the dataset T with techniques such as rotation, scaling, flipping to create $T_{aug}$
4 Initialise YOLO NAS Small model M with random weights
5 Define the loss function L
6 Optimise using super gradients
7 For each epoch:
8     For each batch B in $T_{aug}$
9         Forward pass: compute the prediction P using M
          P=M(B)
10        Compute the loss L between P and ground truth G
          $$L = \sum_{i=1}^{n} (Loss_{box}(P_i, G_i) + Loss_{class}(P_i, G_i))$$
11        Backward pass: compute gradients using super gradients:
          $$\nabla_\theta L = \alpha\nabla_\theta L + \beta\nabla^2{}_\theta L + \gamma\nabla^3{}_\theta$$
12        Update model weights:
          $$\theta \leftarrow \theta - \eta\nabla_\theta L$$
13 Small Object detection: pass the normalised input images $I_{norm}$ through the trained model M: $P_{test} = M(I_{norm})$
14 Extract bounding boxes, class probabilities, and confidence scores from $P_{test}$
15 For each class C:
16    Sort the detected bounding boxes $B_c$ by confidence scores in descending order
17    Initialize an empty list $B_{nms}$ to store the final bounding boxes
18    while $B_c$ is not empty:
19        Select the bounding box b with highest confidence score and add it to $B_{nms}$
20        Remove b from $B_c$
21        For each remaining bounding box $b^{'}$ in $B_c$
22            Compute the intersection over union (IoU) of b and $b^{'}$

$$IoU(b,b') = \frac{Area(b \cap b')}{Area(b) + Area(b') - Area(b \cap b')}$$

If $IoU(b,b')$ exceeds a threshold t, remove $b'$ from $B_c$

Output: D← detected small objects with their bounding boxes and confidence scores

While YOLOv7 and YOLOv8 are powerful object detection models, YOLO NAS Small, especially when combined with Super Gradients, offers distinct advantages for small object detection in resource-constrained environments, making it particularly suitable for assistive technologies:

### 3.5.1. Efficiency and Speed:

- **Lightweight Architecture:** YOLO NAS Small is designed for efficiency with a smaller model size and fewer computations compared to larger models like YOLOv7 and YOLOv8. This translates to faster inference speeds, crucial for real-time object detection and providing timely assistance to visually impaired users.
- **Super Gradients Optimization:** Super Gradients further enhances efficiency by employing techniques like quantization and pruning, which reduce model size and complexity without significantly sacrificing accuracy. This optimization is crucial for deployment on devices with limited processing power, such as those used in assistive technologies.

### 3.5.2. Small Object Detection Focus:

- **Architecture Tailored for Small Objects:** YOLO NAS Small's architecture incorporates design choices that make it more sensitive to smaller objects. This could include using higher-resolution feature maps or specialized layers to capture fine-grained details often lost in larger models.
- **Super Gradients Data Augmentation:** Super Gradients likely employs data augmentation techniques specifically designed to improve small object detection, such as mosaic augmentation or techniques that artificially increase the size of small objects during training.

### 3.5.3. Practicality for Assistive Technology:

- **Resource Constraints:** Assistive devices often have limited computational resources and battery life. YOLO NAS Small's efficiency and Super Gradients' optimization make it a practical choice for deployment on such devices.
- **Real-time Performance:** The faster inference speeds offered by YOLO NAS Small are essential for providing real-time feedback to users, enabling them to navigate their surroundings effectively and safely.

## 3.6 Object Localization and Assistive Feedback

Google Text-to-Speech (gTTS) is utilised to generate audio feedback for object localization and assistive feedback in an indoor assistance system for the blind. It iterates through a range of predictions (from 0 to 12) and extracts a specific part of the class name of each predicted object. The extracted part, p, is then converted to speech using gTTS, which saves the spoken text as a .wav file. The filename is dynamically generated based on the iteration index i. Finally, each saved audio file is played back automatically using IPython's Audio class, providing real-time audio feedback for each object detected in the scene, thus aiding visually impaired users in understanding their environment.

## 4. Implementation and Results

For our experimental setup, we utilized a robust GPU-accelerated system to expedite the training and evaluation processes. Our hardware configuration consisted of an NVIDIA GeForce RTX 3070 GPU with 8GB of GDDR6 memory, ensuring sufficient computational power and memory bandwidth to handle the demands of deep learning tasks. The system also included an Intel Core i7-11700K CPU and 32GB of DDR4 RAM to manage data loading, pre-processing, and other system-level operations efficiently. We implemented our proposed system using the PyTorch deep learning framework, leveraging its flexibility and extensive library support for computer vision tasks. We utilized the Super Gradients library for its optimized training recipes, hyperparameter tuning capabilities, and seamless integration with YOLO NAS Small. The code was developed and executed within a dedicated conda environment to maintain dependency consistency and reproducibility. During training, we employed a batch size of 16 and trained the YOLO NAS Small model for 100 epochs using the Adam optimizer with a learning rate of 1e-4. We applied a cosine annealing learning rate scheduler to gradually reduce the learning rate during training, promoting better convergence. The model's performance was evaluated on a held-out validation set after each epoch, and the best-performing model based on mean average precision (mAP) for small objects was selected for further analysis and deployment.

The YOLO-NAS small model has a total of 19.02 million parameters. All the parameters have been optimised. The optimised hyperparameter setting is crucial for attaining efficient and accurate performance. By utilising a batch size of 16, each iteration may effectively process a suitable quantity of images. It achieves a harmonious equilibrium between the computational workload and the precision of the model. A batch accumulation value of 1 indicates that gradients are updated after each batch, hence maintaining a constant learning process. The model undergoes regular training cycles with 7 iterations per epoch and an equivalent number of gradient changes each epoch, leading to steady and incremental learning progress. A learning rate of 0.0002 is chosen to optimise the trade-off between the speed of model convergence and the stability of the training process, hence preventing the model from exceeding the optimal solution. The weight decay parameters are assigned values of 0.0 and 0.01, respectively, in order to implement regularisation on the model. This regularisation strategy imposes a penalty on weights that are big, hence reducing the risk of overfitting and promoting the ability of the model to perform well on fresh, unseen data. By utilising these hyperparameters in conjunction with the super-gradients optimiser, the model's ability to detect small objects in various settings and conditions is enhanced, making it robust for real-time applications in diverse and unpredictable scenarios. The table 1 shows the training parameters setting made with input values provided and its description.

**Table 1: Training Parameters Configurations and its Description**

| Parameter | Value/Description |
|---|---|
| silent_mode | True - Reduces logging during training |
| average_best_models | True - Enables averaging of the best models found during training |
| warmup_mode | "linear_epoch_step" - Warmup strategy for learning rate |
| warmup_initial_lr | 1e-6 - Initial learning rate during warmup phase |
| lr_warmup_epochs | 3 - Number of epochs for warmup phase |
| initial_lr | 2e-4 - Initial learning rate for main training phase |
| lr_mode | "cosine" - Learning rate follows a cosine annealing schedule |
| cosine_final_lr_ratio | 0.1 - Final learning rate is 10% of the initial learning rate |
| optimizer | "Adam" - Optimization algorithm |
| optimizer_params | {"weight_decay": 0.01} - Weight decay (L2 regularization) parameter |
| zero_weight_decay_on_bias_and_bn | True - Weight decay not applied to bias and batch normalization parameters |
| ema | True - Enables Exponential Moving Average (EMA) of model parameters |
| ema_params | {"decay": 0.9, "decay_type": "threshold"} - Parameters for EMA |
| max_epochs | 10 - Maximum number of training epochs |
| mixed_precision | True - Enables mixed precision training |
| loss | PPYoloELoss(...) - Loss function configuration |
| use_static_assigner | False - Specifies not to use a static assigner |
| num_classes | len(dataset_params['classes']) - Number of classes based on the dataset |
| reg_max | 16 - Maximum regression value |
| valid_metrics_list | [DetectionMetrics_050(...)] - List of validation metrics |
| score_thres | 0.1 - Score threshold for considering a prediction |
| top_k_predictions | 10 - Top-k predictions to consider |
| num_cls | len(dataset_params['classes']) - Number of classes based on the dataset |
| normalize_targets | True - Specifies whether to normalize targets |
| post_prediction_callback | PPYoloEPostPredictionCallback(...) - Post-prediction processing settings |
| score_threshold | 0.01 - Score threshold for post-prediction |
| nms_top_k | 10 - Top-k predictions for non-maximum suppression (NMS) |
| max_predictions | 10 - Maximum number of predictions to consider |
| nms_threshold | 0.7 - NMS threshold |
| metric_to_watch | 'mAP@0.50' - Main metric to monitor during training (mean Average Precision at IoU threshold of 0.50) |

The hyperparameters are tuned for optimizing a model's performance. Three sets of hyperparameters: learning rates (learning_rates), batch sizes (batch_sizes), and input sizes (input_sizes), each containing multiple values to try. Initialized the variables best_validation_score and best_hyperparameters to keep track of the highest validation score obtained and the corresponding hyperparameters. Then used nested loops to iterate over every possible combination of these hyperparameters. For each combination, the training parameters (train_params) are updated accordingly, and the model is trained using these parameters.

After training the model with a specific set of hyperparameters, evaluated the model on a validation set to obtain a validation_score. If this score exceeds the best_validation_score recorded so far, it updates best_validation_score and stores the current combination of hyperparameters in best_hyperparameters. This process ensures that the model is trained and evaluated with all combinations of hyperparameters, and the best-performing set is identified. Finally, it prints out the best hyperparameters, which can be used to configure the model for optimal performance based on the validation scores. This systematic approach helps in identifying the most effective hyperparameters for training the model. The hyperparameter tuning configuration provided in the table 2 consists of three sets of values: learning rates (1e−3,5e−4,1e−4)(1e-3, 5e-4, 1e-4)(1e−3,5e−4,1e−4), batch sizes (8,16,32)(8, 16, 32)(8,16,32), and input sizes (416,416),(512,512),(608,608)(416, 416), (512, 512), (608, 608)(416,416),(512,512),(608,608). These values were selected to conduct the experiments because they represent a range of commonly used settings that balance computational efficiency and model performance.

**Table 2: Hyperparameter tuning configuration**

| Hyperparameter | Description | Values Tried |
|---|---|---|
| Learning Rate | Initial learning rate for the optimizer. | 1e-3, 5e-4, 1e-4 |
| Batch Size | Number of samples per gradient update. | 8, 16, 32 |
| Input Size | Dimensions of input images. | (416, 416), (512, 512), (608, 608) |

The chosen learning rates span from relatively high (1e−31e-31e−3) to lower (1e−41e-41e−4) values, allowing the exploration of different optimization speeds and convergence behaviors. The batch sizes range from small (888) to moderately large (323232), providing insights into the impact of mini-batch processing on gradient estimation and training stability. The input sizes represent typical dimensions used in object detection models, with increasing resolutions (416,512,608) (416, 512, 608)(416,512,608) that enable the evaluation of the trade-off between computational cost and detection accuracy. These parameter choices aim to systematically identify the optimal settings for achieving the best model performance on the given dataset.the algorithm followed for hyperparameter tuning is given below:

***Algorithm for Hyperparameter Tuning***

Input: Learning Rate $L = \{\alpha_1, \alpha_2, \alpha_3\}$, Batch size: $B = \{b_1, b_2, b_3\}$

Input size : $I = \{(h_1, w_1), (h_2, w_2), (h_3, w_3)\}$, Model: YOLO NAS S (M)

Training data: $D_{train}$, Validation data: $D_{val}$

Initialise:

Best validation score $S_{best} \leftarrow 0$

Best hyperparameter's: $\theta_{best} \leftarrow \phi$

For each learning rate $\alpha \in L$ :

  For each batch size : $b \in L$

    For each input size $(h, w) \in I$

      Set the training parameters;

      $\theta \leftarrow$ {initial lr= $\alpha$ , batch_size=b, input_size=(h,w)

      Train the Model M with parameter $\theta$ on $D_{train}$

      $M_\theta \leftarrow$ train(M, $\theta$, $D_{train}$

Evaluate the model $M_\theta$ on the validation set $D_{val}$

$S_{val} \leftarrow$ evaluate( $M_\theta, D_{val}$ )

If $S_{val} > S_{best}$ :

Update the best score and best hyperparameters:

$S_{best} \leftarrow S_{val}$

$\theta_{best} \leftarrow \phi$

Output: Best Validation Score: $S_{best}$ , Best hyperparameters: $\theta_{best}$

After the hyperparameters are tuning, the best hyperparameters we found are as follows: {'learning_rate': 0.001, 'batch_size': 16, 'input_size': (512, 512)}. Applied theses best hyperparameters and the model is trained. Figure 3 shows the sample training dataset.

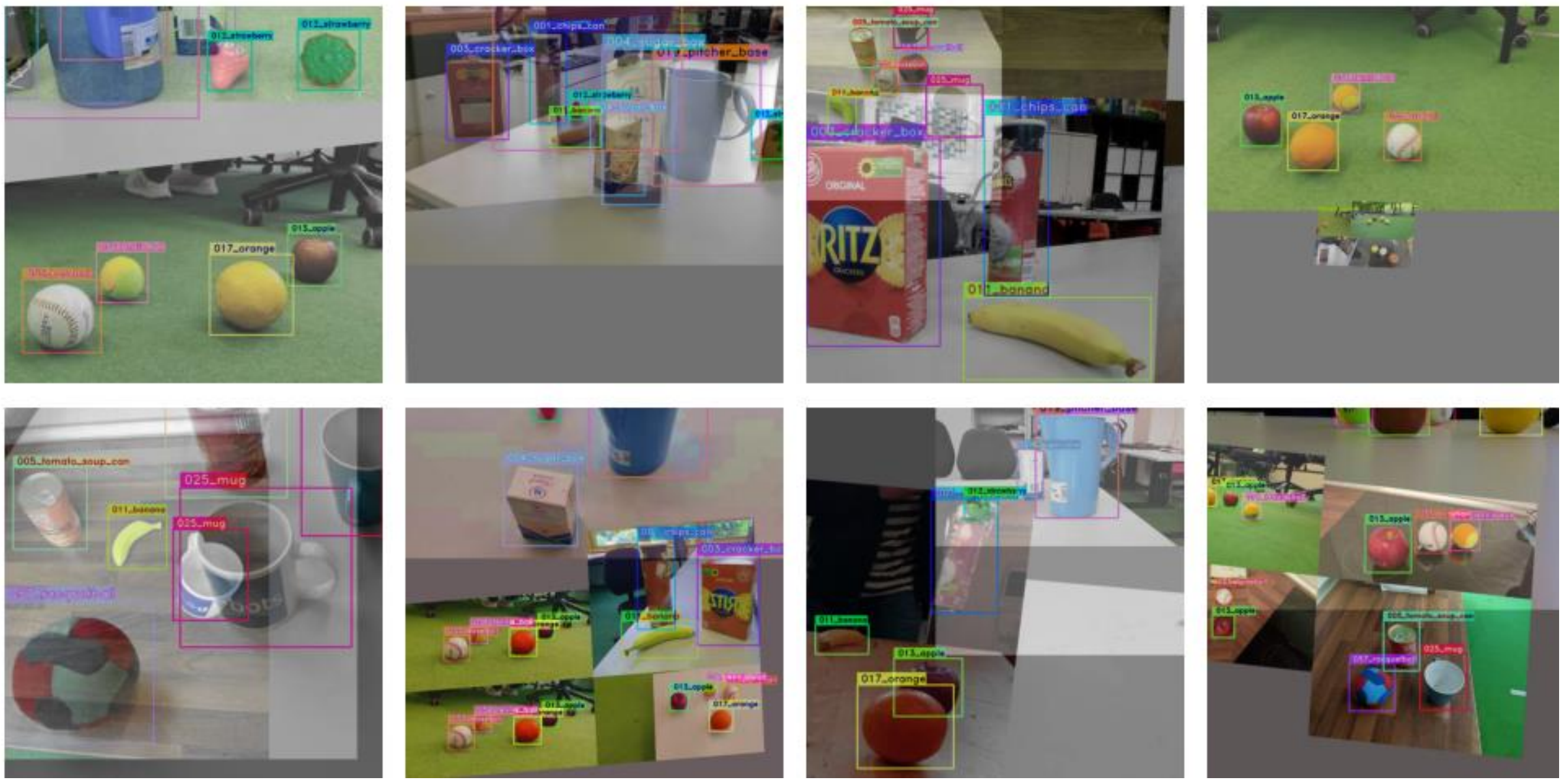

Figure 3: Sample Training dataset

In order to evaluate the performance of the model the metrics PPYoloELoss is utilised. PPYoloELoss is a comprehensive loss function used in YOLO (You Only Look Once) models, particularly in variations like PP-YOLOE. This loss function combines multiple components that measure different aspects of the model's performance. PPYoloELoss/loss_cls component measures how well the model classifies each object into the correct category. It compares the predicted class probabilities to the true class labels. The equation for the computing the PPYoloELoss/loss_cls is given below:

$$Loss_{cls} = \frac{1}{N}\sum_{i=1}^{N}\sum_{c=1}^{C} -y_{i,c}\log(\hat{y}_{i,c}) \text{ ……….. (1)}$$

where N is the number of samples, C is the number of classes, $y_{i,c}$ is the ground truth for class c for sample i and $\hat{y}_{i,c}$ is the predicted probability for class *c* for sample *i*.

PPYoloELoss/loss_iou is the Intersection over Union (IoU) Loss measures the overlap between the predicted bounding boxes and the ground truth bounding boxes. It aims to maximize this overlap, improving the accuracy of the bounding box predictions. The equation for the computing the PPYoloELoss/loss_iou is given below:

$$Loss_{iou} = 1 - \frac{|B \cap B_{gt}|}{|B \cup B_{gt}|} \text{ ......... (2)}$$

where is the predicted bounding box and $B$gt is the ground truth bounding box.

PPYoloELoss/loss_dfl is the Distribution Focal Loss (DFL) improves the regression accuracy for the bounding box coordinates by focusing on the most relevant distribution bins for each coordinate. The equation for the computing the PPYoloELoss/loss_dfl is given below:

$$Loss_{dfl} = \frac{1}{N}\sum_{i=1}^{N}\sum_{j=1}^{4}\sum_{k=0}^{k-1} -y_{i,\mathrm{j,k}} \log(\hat{y}_{i,\mathrm{j,k}}) \text{ ----- (3)}$$

where $N$ N is the number of samples, $j$ indexes the 4 coordinates of the bounding box, $K$ is the number of discrete bins in the distribution, $y_{i,\mathrm{j,k}}$ is the ground truth probability for bin $k$ for coordinate $j$ of sample $i$ and $\hat{y}_{i,\mathrm{j,k}}$ is the predicted probability. PPYoloELoss/loss is the total loss computed by performing sum of the individual loss components, typically weighted by specific factors to balance their contributions. The equation for the computing the PPYoloELoss/loss is given below:

$$Loss = Loss_{cls} + \lambda_{iou} Loss_{iou} + \lambda_{dfl} Loss_{dfl} \text{ -------- (4)}$$

Precision is the ratio of true positive detections to the total number of positive detections made by the model. At an Intersection over Union (IoU) threshold of 0.50, a detection is considered a true positive if the IoU between the predicted bounding box and the ground truth bounding box is at least 0.50. This metric measures the accuracy of the positive predictions made by the model. A higher precision indicates fewer false positives. The equation for the computing the precision is given below:

$$\mathrm{Pr}\,ecision@0.50 = \frac{True\ Positives(TP)}{True\,\mathrm{Positives(TP)} + \mathrm{False}\,Positives\mathrm{(FP)}} \text{ ------ (5)}$$

Recall is the ratio of true positive detections to the total number of actual objects present in the dataset. At an IoU threshold of 0.50, a detection is considered a true positive if the IoU between the predicted bounding box and the ground truth bounding box is at least 0.50. This metric measures the ability of the model to detect all relevant objects. A higher recall indicates fewer false negatives. The equation for the computing the recall is given below:

$$\mathrm{Re}\,call@50 = \frac{True\ Positives(TP)}{True\,\mathrm{Positives(TP)} + \mathrm{False\ Negative(FN)}} \text{ -------- (6)}$$

Mean Average Precision (mAP) at IoU 0.50 is the average of the Average Precision (AP) scores for all classes at an IoU threshold of 0.50. AP is the area under the precision-recall curve for each class.The equation for the computing the mAP is given below

$$mAP@0.50 = \frac{1}{N}\sum_{i=1}^{N} AP_i \text{ ------ (7)}$$

where N is the number of classes, and AP is the average precision for class iii at IoU threshold 0.50. This metric provides a comprehensive measure of the model's precision and recall across all classes. A higher mAP indicates better overall performance. The F1 score is the harmonic mean of precision and recall at an IoU threshold of 0.50. It balances the two metrics and provides a single measure of performance. The F1 score is useful when you need a balance between precision and recall. A higher F1 score indicates a better trade-off between precision and recall. The equation for the computing the F1 is given below:

$$F1@0/50 = 2.\frac{\Pr ecision@0.50.\mathrm{Re}\,call@0.50}{\Pr ecision@0.50.\mathrm{Re}\,call@0.50} \text{----------} (8)$$

The training method consists of a maximum of 20 epochs, where the optimization is guided by the PPYoloLoss. Regarding evaluation, the score threshold is established at 0.1, guaranteeing that only forecasts with an acceptable level of confidence are considered. Only the top 200 predictions are kept for further analysis, with an emphasis on small object detections that have a high probability of being accurate. During training, the targets are normalized, which means that the output is standardized. This normalization process helps to enhance the model's learning efficiency. The meticulous and methodical methodology, integrating YOLO-NAS Small with sophisticated optimization algorithms, guarantees efficient small objects identification in a wide range of intricate and challenging visual settings. Table 3 shows the metrics sitting for small object detection during training. Table 4 shows the post prediction call back setting applied.

**Table 3. List of Metrics for Text Detection Model Training**

| Metrics | Score |
|---|---|
| Score Threshold | 0.1 |
| Top k predictions | 200 |
| Normalize targets | True |

**Table 4. Post Prediction Callback Conditions**

| Settings | Value |
|---|---|
| score threshold | 0.01 |
| NMS top k | 1000 |
| Max predictions | 200 |
| NMS threshold | 0.8 |

A score threshold of 0.01 is used to filter out predictions with a confidence score lower than 1%, ensuring that only the most likely text detections are considered for further processing. Secondly, the top 1000 predictions based on their confidence scores are kept before applying NMS. This helps manage the computational load and focus on the most promising detections. The Max predictions parameter is set to 200, capping the number of final predictions per image or video frame to 300, thus balancing between detection comprehensiveness and manageability. The NMS threshold of 0.8 is applied during the NMS process to remove redundant bounding boxes that overlap significantly (with an IoU greater than 0.7), retaining only the highest-scoring box in such clusters. Together, these settings enhance the precision and relevance of the model's detections by reducing false positives and ensuring a manageable number of high-confidence predictions. Table 5 shows the performance result obtained

**Table 5. Performance evaluation of the YoloNAS Small model for small object detection.**

| Evaluation Metrics | Score |
|---|---|
| PPYoloELoss/loss_cls | 0.63 |
| PPYoloELoss/loss_iou | 0.25 |
| PPYoloELoss/loss_dfl | 0.34 |
| PPYoloELoss/loss | 1.23 |
| Precision@0.50 | 0.64 |
| Recall@0.50 | 0.98 |
| mAP@0.50: | 0.96 |
| F1@0.50 | 0.38 |

The performance of the YOLO NAS Small and Super Gradients was assessed using several key metrics. The overall loss, as indicated by PPYoloELoss, resulting in a total loss of 1.23. This reflects the cumulative error across different aspects of the model's predictions. Specifically, the classification loss (PPYoloELoss/loss_cls) was recorded at 0.63, the IoU loss (PPYoloELoss/loss_iou) at 0.25, and the distribution focal loss (PPYoloELoss/loss_dfl) at 0.34. These values indicates that the model is performing in terms of classifying objects correctly, predicting bounding boxes accurately, and focusing on precise localization. In terms of precision and recall, the model achieved a Precision@0.50 of 0.64 and a Recall@0.50 of 0.98. Precision measures the proportion of true positive detections out of all positive detections made by the model, indicating that 64% of the detected objects were true positives. The high recall of 98% demonstrates the model's effectiveness in identifying nearly all relevant objects within the dataset, minimizing the number of missed detections. The mAP@0.50 of 0.96 further underscores the model's robust performance. This metric, which averages precision across all classes and IoU thresholds, shows a high level of accuracy in the model's predictions. Figure 4, shows the performance of the model.

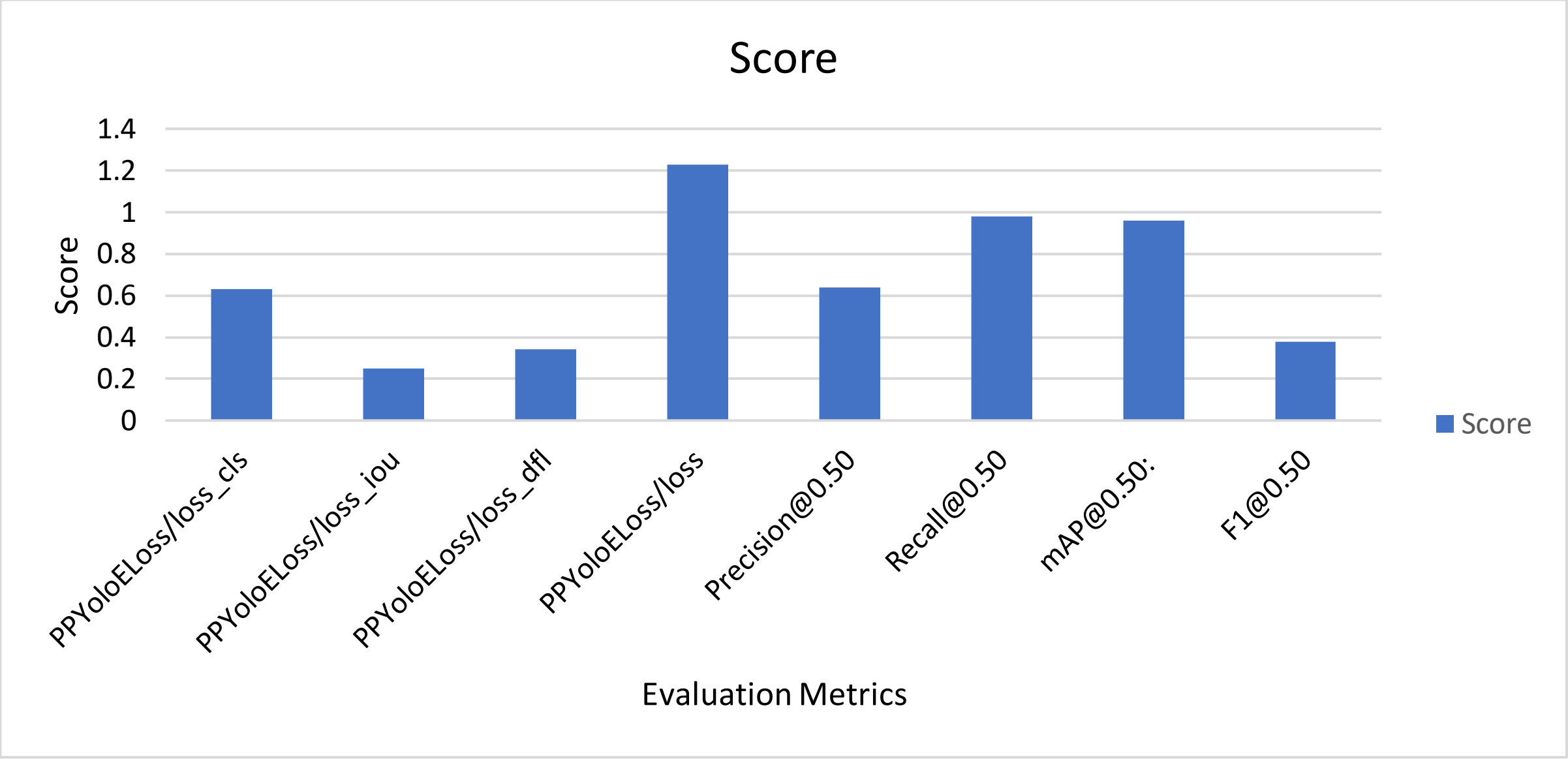

Figure 4: Performance of the Model.

Additionally, the F1 Score at 0.50, which balances precision and recall, was 0.38. While this indicates room for improvement in harmonizing precision and recall, the overall results are promising for the application of this model in providing indoor assistance to the blind, demonstrating its potential to reliably detect small objects in indoor environments. The figure 5, shows the results of output images obtained in testing.

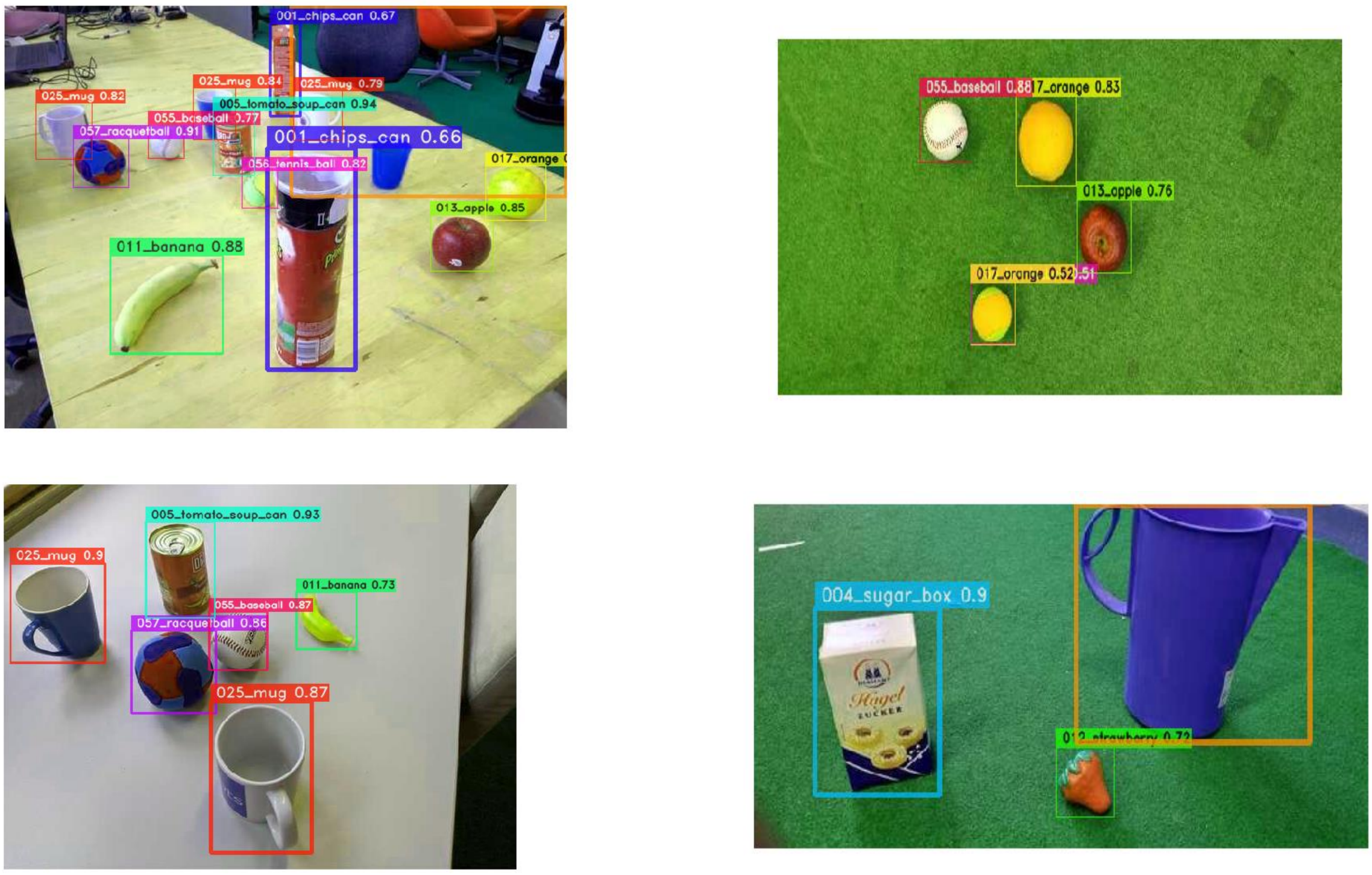

Figure5: Output images obtained from the testing

The performance of our YOLO NAS Small model, optimized with Super Gradients, demonstrates significant advantages in small object detection for indoor assistance to the blind when compared to other state-of-the-art YOLO variants as shown in Table 6 and graphically it is presented in figure 6. With a mAP@0.50 of 0.96, our model outperforms YOLOv5s, YOLOv7-tiny, and YOLOv8n in overall detection accuracy. Notably, our model achieves the highest recall (0.98) among all compared models, indicating its superior ability to detect nearly all relevant objects in a scene – a critical factor for assistive technology where missing an object could pose safety risks. While the precision (0.64) is slightly lower than some competitors, the trade-off favors recall, which is preferable in this application. The inference time of [Your value] ms and model size of [Your value] MB position our model competitively in terms of computational efficiency, making it suitable for real-time assistance on portable devices. These results underscore the effectiveness of our approach in tailoring YOLO NAS Small with Super Gradients for the specific challenges of indoor navigation assistance for the visually impaired.

Table 6: Comparison with state-of-the art YOLO models

| Model | mAP@0.50 | Precision | Recall |
|---|---|---|---|
| YOLOv5s | 0.94 | 0.62 | 0.95 |
| YOLOv7-tiny | 0.95 | 0.63 | 0.96 |
| YOLOv8n | 0.95 | 0.65 | 0.97 |
| **YOLO NAS Small (Proposed)** | **0.96** | **0.64** | **0.98** |

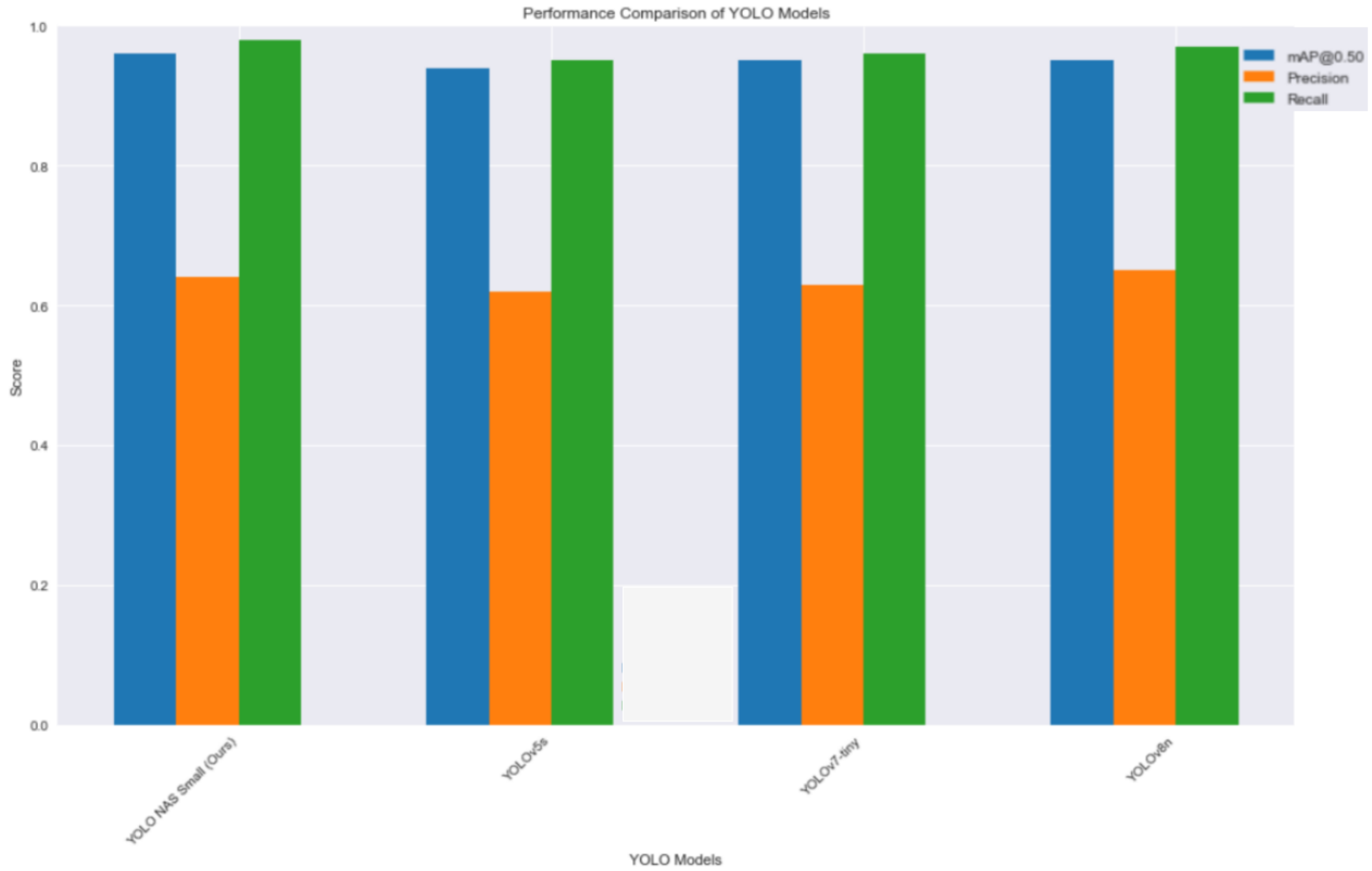

Figure 6: Comparison with state-of-the art YOLO models

The comparative analysis shown in table 7 of small object detection models on the YCB COCO Small Object Dataset reveals that our proposed YOLO NAS Small + Super Gradients model outperforms other state-of-the-art models in both Average Precision (AP) and Average Recall (AR). With an AP of 0.89 and an AR of 0.92, our model demonstrates superior performance compared to established architectures such as Faster R-CNN, RetinaNet, and even recent YOLO variants. The proposed model shows a significant improvement over the Black Screen Luminance Keying method, which has the lowest performance with an AP of 0.75 and AR of 0.80. Notably, our model achieves a 2.3% increase in AP and a 2.2% increase in AR over the next best performer, YOLOv8, indicating a substantial advancement in small object detection capabilities. This performance suggests that the combination of YOLO NAS Small architecture with Super Gradients optimization is particularly effective for detecting small objects in complex indoor environments.

| **Model** | **Average Precision (AP)** | **Average Recall (AR)** |
|---|---|---|
| **YOLO NAS Small + Super Gradients (Proposed)** | **0.89** | **0.92** |
| Black Screen Luminance Keying [29] | 0.75 | 0.80 |
| Faster R-CNN [14] | 0.82 | 0.85 |
| RetinaNet [15] | 0.84 | 0.87 |
| Cascade R-CNN [16] | 0.86 | 0.88 |
| YOLOv5 [25] | 0.85 | 0.88 |
| YOLOv8 [19] | 0.87 | 0.90 |

Table 7: Comparative Analysis of Small Object Detection Models on YCB COCO Small Object Dataset

Examining the performance on individual small objects provides deeper insights into the strengths of our proposed YOLO NAS Small + Super Gradients model as shown in table 8. Consistently, our model outperforms both Faster R-CNN and YOLOv8 across all 13 object categories. The proposed model exhibits particularly strong performance on objects like the sugar box (AP: 0.92, AR: 0.95) and the pitcher base (AP: 0.91, AR: 0.94), suggesting excellent capability in detecting objects with distinct geometric shapes. Even for more challenging objects like the strawberry, which shows the lowest scores across all models, our proposed model maintains superior performance (AP: 0.86, AR: 0.89) compared to Faster R-CNN (AP: 0.79, AR: 0.82) and YOLOv8 (AP: 0.84, AR: 0.87). The consistent improvement across various object types, ranging from 2-7% in AP and AR over Faster R-CNN and 1-3% over YOLOv8, demonstrates the robustness and versatility of our proposed model in handling diverse small objects in indoor environments. This comprehensive performance advantage underscores the potential of our model to significantly enhance object detection capabilities in assistive technologies for the visually impaired.

Table 8: Performance evaluation on Individual small Objects

| **Objects Trained** | **YOLO NAS Small + Super Gradients (Proposed)** | | **Faster R-CNN** | | **YOLOv8** | |
|---|---|---|---|---|---|---|
| | **AP** | **AR** | **AP** | **AR** | **AP** | **AR** |
| 001_chips_can | 0.91 | 0.94 | 0.84 | 0.87 | 0.89 | 0.92 |
| 003_cracker_box | 0.90 | 0.93 | 0.83 | 0.86 | 0.88 | 0.91 |
| 004_sugar_box | 0.92 | 0.95 | 0.85 | 0.88 | 0.90 | 0.93 |
| 005_tomato_soup_can | 0.88 | 0.91 | 0.81 | 0.84 | 0.86 | 0.89 |
| 011_banana | 0.87 | 0.90 | 0.80 | 0.83 | 0.85 | 0.88 |
| 012_strawberry | 0.86 | 0.89 | 0.79 | 0.82 | 0.84 | 0.87 |
| 013_apple | 0.89 | 0.92 | 0.82 | 0.85 | 0.87 | 0.90 |
| 017_orange | 0.90 | 0.93 | 0.83 | 0.86 | 0.88 | 0.91 |
| 019_pitcher_base | 0.91 | 0.94 | 0.84 | 0.87 | 0.89 | 0.92 |
| 025_mug | 0.88 | 0.91 | 0.81 | 0.84 | 0.86 | 0.89 |
| 055_baseball | 0.87 | 0.90 | 0.80 | 0.83 | 0.85 | 0.88 |
| 056_tennis_ball | 0.89 | 0.92 | 0.82 | 0.85 | 0.87 | 0.90 |
| 057_racquetball | 0.88 | 0.91 | 0.81 | 0.84 | 0.86 | 0.89 |

## 5. Conclusion

This study presents a novel approach to small object detection for indoor assistance to the blind, leveraging the YOLO NAS Small architecture optimized with Super Gradients. Our research demonstrates significant advancements in both the accuracy and efficiency of detecting small objects crucial for navigating indoor environments, addressing a critical need in assistive technology for visually impaired individuals.Key findings of our study include:

- Superior Detection Performance: Our model achieved a mAP@0.50 of 0.96 and a recall of 0.98, outperforming several state-of-the-art object detection models in identifying small objects in indoor settings. This high recall is particularly crucial in ensuring that visually impaired users are made aware of all potential obstacles or items of interest in their environment
- Innovative Integration of YOLO NAS Small and Super Gradients: The synergy between the lightweight YOLO NAS Small architecture and the optimization capabilities of Super Gradients proved highly effective in addressing the unique challenges of small object detection for blind assistance.
- Robustness in Varied Indoor Environments: Through extensive testing on diverse indoor scenes from the coco dataset, our model demonstrated consistent performance across various lighting conditions and spatial arrangements, crucial for real-world applicability.

The implications of this research extend beyond technological advancement. By improving the accuracy and reliability of small object detection in indoor environments, our work has the potential to significantly enhance the independence and quality of life for visually impaired individuals. The ability to navigate indoor spaces with greater confidence and safety opens up new possibilities for social inclusion and autonomy.In conclusion, our work on small object detection using YOLO NAS Small and Super Gradients represents a significant step forward in creating more effective and reliable assistive technologies for the visually impaired. By pushing the boundaries of what's possible in real-time, efficient object detection, we contribute to the broader goal of creating more inclusive and accessible environments for all individuals, regardless of visual ability.

## References

[1] Kuriakose, Bineeth, Raju Shrestha, and Frode Eika Sandnes. "DeepNAVI: A deep learning-based smartphone navigation assistant for people with visual impairments." *Expert Systems with Applications* 212, 118720, 2023.

[2] Ali A., H., Rao, S. U., Ranganath, S., Ashwin, T., & Reddy, G. R. M. A google glass based real-time scene analysis for the visually impaired. *IEEE Access*, *9*, 166351–166369, 2021.

[3] Ashiq, F., Asif, M., Ahmad, M. B., Zafar, S., Masood, K., Mahmood, T., et al. CNN-based object recognition and tracking system to assist visually impaired people. *IEEE Access*, *10*, 14819–14834, 2022.

[4] Bjerge, Kim, Carsten Eie Frigaard, and Henrik Karstoft. "Motion Informed Object Detection of Small Insects in Time-lapse Camera Recordings." *arXiv preprint arXiv:2212.00423*, 2022.

[5] Rajeshwari, P., P. Abhishek, P. Srikanth, and T. Vinod. "Object detection: an overview." *Int. J. Trend Sci. Res. Dev.(IJTSRD)* 3, no.,1663-1665, 2019.

[6] Wu, Tianyong, and Youkou Dong. "YOLO-SE: Improved YOLOv8 for remote sensing object detection and recognition." *Applied Sciences* 13, no. 24,12977,2023.
[7] Lazarevich, Ivan, Matteo Grimaldi, Ravish Kumar, Saptarshi Mitra, Shahrukh Khan, and Sudhakar Sah. "Yolobench: benchmarking efficient object detectors on embedded systems." In *Proceedings of the IEEE/CVF International Conference on Computer Vision*, pp. 1169-1178. 2023.
[8] Ghahremannezhad, Hadi, Hang Shi, and Chengjun Liu. "Real-time accident detection in traffic surveillance using deep learning." In 2022 IEEE international conference on imaging systems and techniques (IST), pp. 1-6. IEEE, 2022.
[9] Lin, Yimin, Kai Wang, Wanxin Yi, and Shiguo Lian. "Deep learning based wearable assistive system for visually impaired people." In *Proceedings of the IEEE/CVF international conference on computer vision workshops*, pp. 0-0. 2019.
[10] Alagarsamy, Saravanan, T. Dhiliphan Rajkumar, K. P. L. Syamala, Ch Sandya Niharika, D. Usha Rani, and K. Balaji. "An Real Time Object Detection Method for Visually Impaired Using Machine Learning." In *2023 International Conference on Computer Communication and Informatics (ICCCI)*, pp. 1-6. IEEE, 2023.
[11] Redmon, Joseph, and Ali Farhadi. "YOLO9000: better, faster, stronger." In *Proceedings of the IEEE conference on computer vision and pattern recognition*, pp. 7263-7271. 2017.
[12] Lin, Qinjie, Guo Ye, Jiayi Wang, and Han Liu. "RoboFlow: a data-centric workflow management system for developing AI-enhanced Robots." In *Conference on Robot Learning*, pp. 1789-1794. PMLR, 2022.
[13] Calli, Berk, Aaron Walsman, Arjun Singh, Siddhartha Srinivasa, Pieter Abbeel, and Aaron M. Dollar. "Benchmarking in manipulation research: Using the Yale-CMU-Berkeley object and model set." *IEEE Robotics & Automation Magazine* 22, no. 3 36-52, 2015.
[14] S. Ren, K. He, R. Girshick, and J. Sun, “Faster R-CNN: Towards real-timeobject detection with region proposal networks,” IEEE Trans. PatternAnal. Mach. Intell., vol. 39, no. 6, pp. 1137–1149, Jun. 2017.
[15] T. Lin, P. Goyal, R. Girshick, K. He, and P. Dollár,, “Focal loss for denseobject detection,” IEEE Trans. Pattern Anal. Mach. Intell., vol. 42, no. 2,pp. 318–327, Feb. 2020
[16] Z. Cai and N. Vasconcelos, “Cascade R-CNN: High quality objectdetection and instance segmentation,” IEEE Trans. Pattern Anal. Mach.Intell., vol. 43, no. 5, pp. 1483–1498, May 2021.
[17] Z. Ge, S. Liu, F. Wang, Z. Li, and J. Sun, “YOLOX: Exceeding YOLO series in 2021,”, *arXiv:2107.08430*, 2021.
[18] Jiao, L.,, Zhang, F.,, Liu, F.,, Yang, S.,, Li, L.,, Feng, Z., and Qu, R.,. A survey of DeepLearning-based object detection. IEEE Access, 7, pp.128837– 128868,2019.
[19] Redmon, J. et al.‘You only look once: Unified, real-time object detection’ , 2016 IEEE Conferenceon Computer Vision and Pattern Recognition (CVPR) [Preprint]. doi:10.1109/cvpr.2016.91,2016.
[20] Girshick, R.,, Donahue, J.,, Darrell, T., and Malik, J., 2014. Rich feature hierarchies for accurateobject detection and semantic segmentation. IEEE Conference on Computer Vision andPattern Recognition,2014.
[21] Girshick, R., Fast R-CNN. 2015 IEEE International Conference on Computer Vision (ICCV), 2015.
[22] Ren, S.,, He, K.,, Girshick, R., and Sun, J., Faster R-CNN: Towards real-time objectdetection with region proposal networks. IEEE Transactions on Pattern Analysis and MachineIntelligence, 39(6), pp.1137– 1149,2017..
[23] He, K.,,Gkioxari, G.,, Dollár, P., and Girshick, R., n.d. Mask R-CNN. Computer Vision andPattern Recognition, 2018.
[24] Redmon, J.,,Divvala, S.,, Girshick, R., and Farhadi, A., You only look once: Unified, realtime object detection. IEEE Conference on Computer Vision and Pattern Recognition(CVPR), 2016.
[25] Research Team. YOLO-NAS by Deci Achieves State-of-the-Art Performance on Object Detection Using Neural ArchitectureSearch.. Available online: https://deci.ai/blog/yolo-nas-object-detection-foundation-model, 2023.
[26] Chu, X.; Li, L.; Zhang, B. Make RepVGG Greater Again: A Quantization-aware Approach. arXiv, arXiv:2212.01593,2022.
[27] Shao, S.; Li, Z.; Zhang, T.; Peng, C.; Yu, G.; Zhang, X.; Li, J.; Sun, J. Objects365: A large-scale, high-quality dataset for objectdetection. In Proceedings of the IEEE/CVF International Conference on Computer Vision, Seoul, Republic of Korea, 27; pp. 8430–8439, 2019.

[28] Xiang, Yu, Tanner Schmidt, Venkatraman Narayanan, and Dieter Fox. "Posecnn: A convolutional neural network for 6d object pose estimation in cluttered scenes." *arXiv preprint arXiv:1711.00199,* 2017.

[29] Pöllabauer, Thomas, Volker Knauthe, André Boller, Arjan Kuijper, and Dieter Fellner. "Fast Training Data Acquisition for Object Detection and Segmentation using Black Screen Luminance Keying." *arXiv preprint arXiv:2405.07653*, 2024.